\documentclass[11pt,a4paper]{article}
\usepackage{times,latexsym}
\usepackage{url,hyperref}
\usepackage[T1]{fontenc}
\usepackage{tabularray}
\usepackage{array}
\usepackage{booktabs}
\usepackage{svg}
\usepackage{listings}

\usepackage[acceptedWithA]{tacl2021v1}
\usepackage{xspace,mfirstuc,tabulary}

\newif\iftaclinstructions
\taclinstructionsfalse %
\iftaclinstructions
\renewcommand{\confidential}{}
\renewcommand{\anonsubtext}{(No author info supplied here, for consistency with
TACL-submission anonymization requirements)}
\newcommand{\instr}
\fi

\iftaclpubformat %

\else

\fi

\usepackage{graphicx} %
\usepackage[T1]{fontenc}
\usepackage{babel}
\usepackage{amsmath}
\usepackage{amssymb}

\usepackage[normalem]{ulem}
\usepackage{booktabs}

\newcommand{\choice}[0]{{\textsc{Choice}}}
\newcommand{\likert}[0]{{\textsc{Likert}}}
\newcommand{\logprobs}[0]{{\textsc{LogProbs}}}
\newcommand{\ewok}[0]{{\textsc{EWoK}}}
\newcommand{\ewokone}[0]{{\ewok\textsc{-core-1.0}}}

\title{Elements of World Knowledge (\textsc{EWoK}): \\ A Cognition-Inspired Framework for Evaluating \\ Basic World Knowledge in Language Models%
}

\author{
Anna A.~Ivanova$^{*1}$ %
\And Aalok Sathe$^{*2}$
\And Benjamin Lipkin$^{*3}$ %
\AND 
Unnathi U.~Kumar$^1$ \And Setayesh Radkani$^3$ \And Thomas H.~Clark$^3$ \And Carina Kauf$^3$
\AND Jennifer Hu$^4$ \And R.~T. Pramod$^3$ \And Gabriel Grand$^3$ \And Vivian C. Paulun$^3$ \AND
Maria Ryskina$^5$ \And Ekin Akyürek$^3$ \And Ethan G. Wilcox$^6$ \And Nafisa Rashid$^7$ \And Leshem Choshen$^{3,8}$ 
\AND Roger Levy$^3$ \And Evelina Fedorenko$^3$ \And Joshua Tenenbaum$^3$  \And Jacob
Andreas$^3$ 
\AND 
\normalfont 
$^1$Georgia Tech \\
\normalfont 
$^4$Johns Hopkins University\\ 
\normalfont 
$^7$UC Berkeley \And  
\normalfont 
$^2$Brown University \\ 
\normalfont 
$^5$Vector Institute \\ 
\normalfont 
$^8$MIT-IBM Watson AI 
\And
\normalfont 
$^3$MIT\\
\normalfont 
$^6$Georgetown University
\AND 
\normalfont{\small\ttfamily a.ivanova@gatech.edu, aalok@brown.edu, lipkinb@mit.edu}
\\\normalfont{\small$^*$authors contributed equally}
}

\begin{document}
\maketitle

\begin{abstract}
The ability to build and 
reason about models of the world is essential for situated language understanding.
But evaluating world modeling capabilities in modern AI systems---especially those based on language models---has proven challenging, in large part because of the difficulty of disentangling \emph{conceptual} knowledge about the world from knowledge of surface co-occurrence statistics.
This paper presents Elements of World Knowledge 
(\ewok), 
a framework for evaluating language models' understanding of the conceptual knowledge underlying world modeling. %
\ewok\ targets specific concepts from multiple knowledge domains known to be important for world modeling in humans, from social interactions (\emph{help, deceive}) to spatial relations (\emph{left, right}). 
Objects, agents, and locations in the items can be flexibly filled in, enabling easy generation of multiple controlled datasets. 
We then introduce \ewokone, a dataset of 
4,374 %
items covering $11$ world knowledge domains.
We evaluate 
$20$ open-weights large language models 
($1.3$B--$70$B parameters) and compare them with human performance. 
All tested models perform worse than humans, with results varying drastically across domains. Performance on social interactions and social properties was highest and performance on physical relations and spatial relations was lowest.
Overall, this dataset highlights simple cases where even large models struggle and presents rich avenues for targeted research on LLM world modeling capabilities.

\end{abstract}

\section{Introduction}

\begin{figure*}[tb!]
  \centering
  \includegraphics[width=\textwidth]{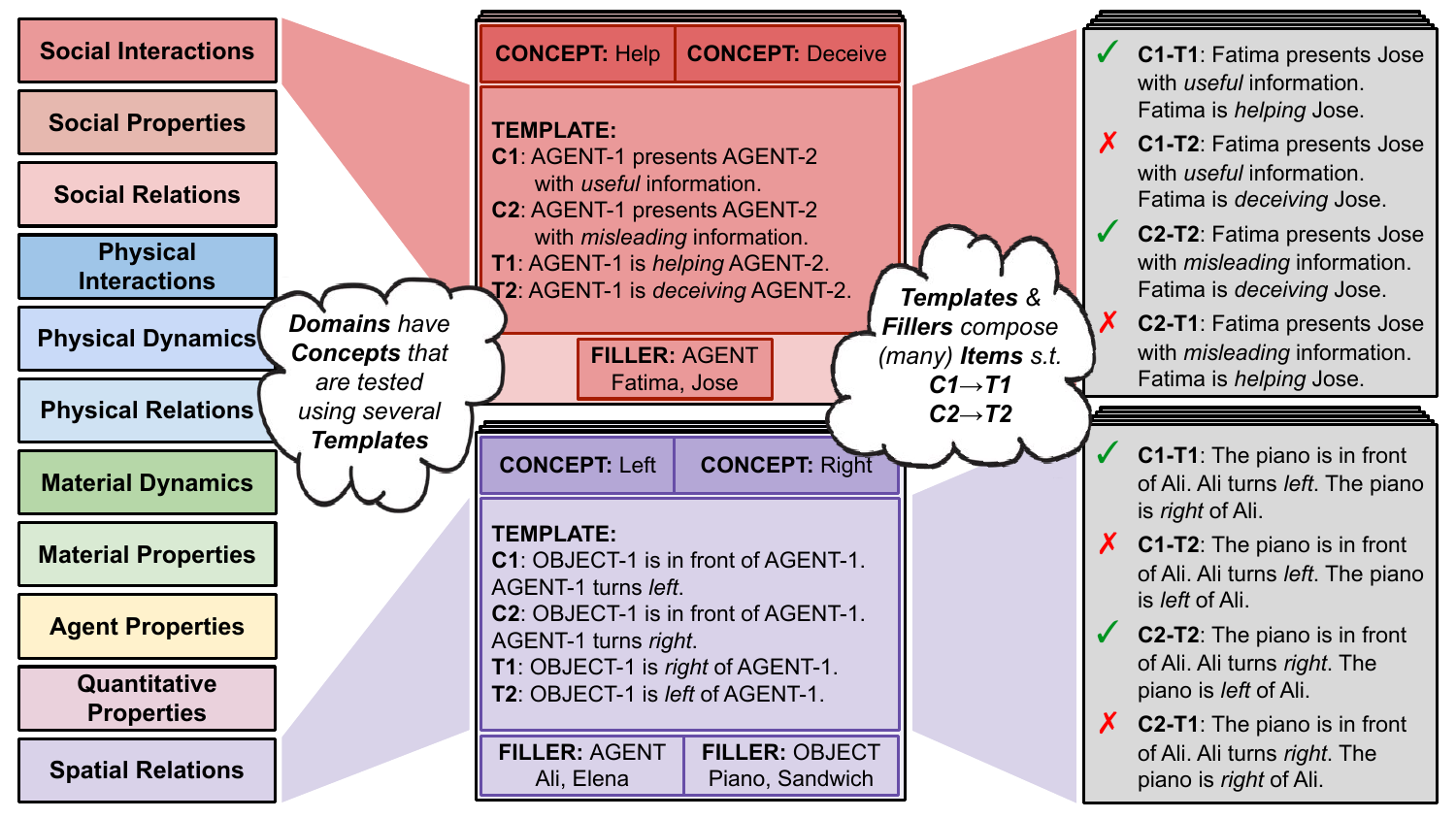}
  \caption{\ewok\ design, illustrated with examples from \emph{social interactions} \& \emph{spatial relations}.
  Each \emph{domain} contains a set of \emph{concepts}, \emph{contexts}, and \emph{targets}. These combine to form many \emph{templates}, which specify minimal pairs of contexts ($C$) and targets ($T$), such that $T_1$ matches $C_1$ but not $C_2$, and $T_2$ matches $C_2$ but not $C_1$. 
  Each template can be combined with \emph{fillers} to generate an even larger collection of \emph{items}.}
  \label{fig:design}
\end{figure*}

Large language models (LLMs) acquire a substantial amount of knowledge from their training data \citep{bender2020climbing,grand2022semantic,pavlick2022semantic}.
This knowledge comprises both \emph{knowledge about language} (e.g.\ word meanings and rules of syntax) and \emph{knowledge about the world} (e.g.\ social conventions and physical properties of objects).
Contemporary LLMs demonstrate substantial knowledge of language \citep{mahowald2024dissociating}, as evidenced by the fluency of the text that they generate. But how robust is their understanding of the basic social, physical, and relational concepts that are foundational to our everyday experience? 

In this paper, we present Elements of World Knowledge (\ewok)\footnote{Data and associated code can be accessed at 
 \url{http://ewok-core.github.io}
}, a flexible framework for evaluating world modeling in LLMs (Figure~\ref{fig:design}). 
The \ewok\ framework consists of: 
\textbf{(a)} several {\itshape domains} that constitute the foundation for basic human world knowledge and are processed by dedicated cognitive and neural systems;
\textbf{(b)} a set of \emph{concepts} within each domain; 
\textbf{(c)} a set of \emph{templates} (and modular components to procedurally generate these templates) that test knowledge of specific concepts by contrasting plausible and implausible context--target combinations; \textbf{(d)} a set of \emph{fillers} to populate the templates, such that each template can be used multiple times; \textbf{(e)} a pipeline to generate a specific set of \emph{items} (a dataset) based on these source materials; and \textbf{(f)} an \emph{evaluation procedure} that measures language models' ability to distinguish plausible and implausible items.

\paragraph{Why elements?} Our framework targets specific cognitive concepts (or concept pairs, such as \emph{left}/\emph{right}). 
Concept knowledge is not limited to memorized definitions, but can be used productively across a wide range of scenarios.
Thus concepts (not isolated sentences or facts) are the primary focus of the \ewok\ framework.
This approach stands in contrast to many NLP datasets, which often include naturalistic, lengthy stimuli that draw on multiple unspecified concepts. Although naturalistic tasks are useful for evaluating model performance in real-life settings, individual item complexity and lack of controls can make it difficult to assess why a model fails. For instance, LLMs exhibit mixed performance on theory of mind tasks, which has led to disagreements about whether LLMs fail at inferring mental states or auxiliary cognitive abilities, such as knowing what it means for a box to be ``transparent'' \citep{kosinski2023theory,ullman2023large}. Our framework mitigates this problem by explicitly linking individual items with the specific concepts that they test. 

\paragraph{Why cognition-inspired?} 
\label{sec:para:cog-inspired}
World knowledge is a notoriously fuzzy capability; which knowledge domains should one focus on? Dataset design in NLP is often centered around data availability, with examples sourced from video captions, Wikipedia articles, or social media logs. In contrast, here, we systematically select a range of knowledge domains that have been shown to recruit dedicated cognitive and/or neural machinery in humans, such as knowledge of intuitive physics \citep{mccloskey1983intuitive, battaglia2013simulation}, knowledge of physical and spatial relations \citep{hafri2021perception}, intuitive number sense \citep{dehaene2011number}, social reasoning \citep{carlson2013theory,thomas2024cognitive}, and reasoning about agents that involves both physical and social knowledge \citep{liu_outa_akbiyik_2024}. These knowledge domains are not specific to language \citep{jackendoff2002foundations}; in fact, many are present in preverbal infants \citep{spelke2007core}. At the same time, traces of these signals are deeply embedded in our daily communicated language. Text itself thus contains rich information that reflects grounded world knowledge \citep{roads2020learning, abdou2021can, patel2021mapping}, and it may be reasonable to predict that LLMs would acquire this domain-specific knowledge from text alone.

\paragraph{Why plausibility?} Concept understanding is often fuzzy. Thus, deciding whether a factual statement is true or false in isolation is often ill-defined. Instead, we use combinations of plausible vs.~implausible context--target pairs. In \ewok, a model needs to determine which scenarios make more sense (more plausible). Having an accurate world model would enable a model to consistently distinguish plausible and implausible scenarios no matter how they are worded.

\paragraph{Why minimal pairs (of pairs)?} Both contexts and targets in \ewok\ have a minimal-pairs design, such that a specific targeted change to a sentence (e.g., \emph{left} $ \rightarrow$ \emph{right}) yields an opposite effect (e.g., \emph{plausible} $\rightarrow$ \emph{implausible}). This approach can help identify specific manipulations that LLMs are and are not sensitive to, with the goal of targeted diagnostics.
Such controlled manipulations are also particularly well-suited for mechanistic interpretability research, which may seek to explain the circuits through which models deploy world knowledge in context.

\paragraph{Why context--target combinations?} LLMs have a remarkable capacity for memorization, such that many plausible and implausible sentences can be distinguished solely based on their presence in the training data (e.g., \emph{The fox chased the rabbit} is more common than \emph{The rabbit chased the fox}). In contrast, our framework tests LLMs' ability to evaluate contextual plausibility, such that the same exact target (\emph{The piano is left of Ali}) is either plausible or implausible depending on the context (see Figure \ref{fig:design} right).
Establishing this control is critical for isolating contextually-sensitive world knowledge from heuristic knowledge associated with surface forms.

\vspace{0.6em}

\noindent Our paper is structured as follows. In Section \ref{sec:related-work}, we review prior research related to the topics explored in this work. In Section \ref{sec:framework}, we describe the core components of the \ewok\ framework. In Section \ref{sec:evaluation}, we describe our evaluation strategy. In Section \ref{sec:results}, we show LLM performance on \ewokone, a dataset generated via the \ewok\ framework. In Section \ref{sec:discussion}, we discuss our results and future prospects for this line of work. Finally, in Section \ref{sec:release}, we discuss dataset release considerations.

\renewcommand{\arraystretch}{0.85} %
\begin{table*}[]
  \footnotesize %

    \centering
    \begin{tabular}{
    llll
    }
    \toprule
        \bf Domain & \bf  Templates & \bf Concepts & \bf Example 
\\
\midrule
   Social Interactions & 165 & 16  & \begin{tabular}{@{}l@{}} C: AGENT-1 presents AGENT-2 with [useful/misleading] information. \\ T: AGENT-1 is \textbf{[helping/deceiving]} AGENT-2.\end{tabular} 
 \\ 
 \midrule
   Social Properties & 185 & 16 &
   \begin{tabular}{@{}l@{}} C: {AGENT-1} [can/cannot] be depended upon. \\ T: {AGENT-1} is \textbf{[trustworthy/untrustworthy]}.
\end{tabular}\\ 
 \midrule
   Social Relations & 785 & 15 & \begin{tabular}{@{}l@{}} C: {AGENT-[1/2]} lectures {AGENT-[2/1]}. \\T: {AGENT-1} is AGENT-2’s \textbf{[teacher/student]}.
\end{tabular}\\  
 \midrule
   Physical Interactions & 280 & 20 & \begin{tabular}{@{}l@{}} C: {OBJECT-1} is moving [toward/away from] the ground. \\ T: {AGENT-1} \textbf{[dropped/lifted]} {OBJECT-1}. 
   \end{tabular}\\  
 \midrule
   Physical Dynamics & 75 & 12 & \begin{tabular}{@{}l@{}} C: The speck on OBJECT-1 [is/is not] rotating. \\ T: OBJECT-1 is \textbf{[rolling/sliding]}.
\end{tabular}\\  
 \midrule
   Physical Relations & 435 & 20 & \begin{tabular}{@{}l@{}} C: OBJECT-1 occupies [more/less] space than OBJECT-2. \\ T: {OBJECT-[1/2]} is \textbf{bigger} than {OBJECT-[2/1]}.
\end{tabular}\\  
 \midrule
   Material Dynamics & 780 & 21 & \begin{tabular}{@{}l@{}} C: {AGENT-1} sees something that is [fabric/liquid]. \\ T: {AGENT-1} can \textbf{[fold/pour]} it.
\end{tabular}\\ 
 \midrule
   Material Properties & 125 & 18 & \begin{tabular}{@{}l@{}} C: {AGENT-1} [can/cannot] see through {OBJECT-1}. \\ T: {OBJECT-1} is \textbf{[transparent/opaque]}.
\end{tabular}\\ 
 \midrule
   Agent Properties & 1130 & 22 & \begin{tabular}{@{}l@{}} C: {AGENT-1} likes {OBJECT-1} [more/less] than {OBJECT-2}.\\ T: {AGENT-1} \textbf{prefers} {OBJECT-[2/1]} to {OBJECT-[1/2]}.
\end{tabular}\\ 
 \midrule
   Quantitative Properties & 195 & 18 &\begin{tabular}{@{}l@{}} C: {AGENT-1} [needs/does not need] more {OBJECT-1}. \\ T: {AGENT-1} has \textbf{[not enough/enough]} {OBJECT-1}.
\end{tabular}\\ 
 \midrule
   Spatial Relations & 245 & 14 & \begin{tabular}{@{}l@{}} C: OBJECT-1 is in front of AGENT-1. AGENT-1 turns [left/right]. \\ T: OBJECT-1 is to the \textbf{[right/left]} of AGENT-2. \end{tabular}
        \\
        \bottomrule
    \end{tabular}
    \caption{\ewokone\ contains 11 domains, each contributing between $75$ and $1130$ templates testing between $12$ and $22$ concepts. Here we include sample templates (pairs of context--target pairs) for each domain. Each target includes an explicit mention of a concept from that domain (highlighted in bold).
    }
    \label{tab:domains}
\end{table*}

\section{Related Work} \label{sec:related-work}

The capabilities we evaluate are closely related to the notion of commonsense knowledge, an area that has numerous text-based benchmarks \citep[e.g.,][]{levesque2012winograd, sakaguchi2021winogrande, zellers2019hellaswag}, including targeted evaluations of physical \citep{bisk2020piqa} and social commonsense \citep{sap2019social}. LLMs often struggle on such commonsense benchmarks, likely due to the reporting bias in their training data \citep{shwartz2020neural}: conversations and texts typically do not include commonly observed or obvious information \citep{gordon2013reporting}. Thus, although our items target basic world knowledge and are not designed to be challenging, it is possible that LLMs could still struggle with them due to the reporting bias. 

One specific version of reporting bias affects perceptually grounded knowledge. Co-occurrence information that is easily available through perception (e.g., the fact that bananas are typically yellow or wheels are typically round) is often underrepresented in language corpora. This bias has led an earlier generation of language models to underperform on physically and perceptually grounded world knowledge tasks \citep{lucy2017distributional,utsumi2020exploring} and exhibit representational differences in physical features that are more/less talked about \citep{abdou2021can, lewis2019distributional}. That said, in spite of such biases, models trained on text do learn a substantial amount of distributional information from perceptual domains \citep{roads2020learning, abdou2021can,sorscher2022neural}, meaning that much of the perceptual world knowledge can be acquired without grounding. Overall, we expect LLMs to perform above-chance on world knowledge domains that are perceptually grounded (such as physical relations or material properties), although they might be more challenging than, e.g., social domains.

Our evaluation is also related to works on natural language inference and entailment. 
The recognizing textual entailment (RTE) task \cite{dagan2010recognizing} poses two sentences to a system (a text expression ${T}$ and hypothesis $H$) and asks that it determine whether $H$ follows from $T$.
The natural language inference (NLI) task follows a similar challenge
\citep{bowman2015large, williams2018broad, conneau2018xnli} and involves making a 3-way judgment about whether a \textit{premise} entails, contradicts, or is neutral relative to a \textit{hypothesis}. \ewok\ asks whether a target sentence $T$ is plausible given a context $C$, which might---but does not have to---indicate an entailment relationship between the two. Though widely successful as a challenge, a large body of subsequent work has highlighted issues with RTE- and NLI-style evaluation: language models can often use heuristics (such as artifacts left behind by human annotators and lexical statistics) to ``solve'' the task without meaningful semantic understanding or reasoning \citep{poliak2018hypothesis,liu2020hyponli,mccoy2020right,gururangan2018annotation}. We address this limitation by (1) posing the task as a minimal pair, where each of two targets is held static while paired contexts are used to modulate preferences, rendering it impossible to rely solely on target plausibility; (2) annotating minimal pair contrast type to test whether any item design feature drives model performance; and (3) testing the relationship between LLM performance and surface-level item properties, such as item length, average word frequency, and performance of a baseline bag-of-words embedding model.

Our approach to dataset design is similar in spirit to the bAbi framework \citep{babi}, which used simple synthetic tasks probing world knowledge and reasoning; however, our items are both simpler in design (they target individual concepts and do not require multi-chain reasoning) and are harder in practice (a minimal pair, context-dependent design greatly reduces the availability of response heuristics, a serious problem in bAbi; \citealp{kaushik2018much}). 

The minimal pair design is common in datasets inspired by psycholinguistics and cognitive science, such as SyntaxGym \citep{gauthier2020syntaxgym}, BLiMP \citep{warstadt2020blimp}, and COMPS \citep{misra2023comps}. In particular, it has previously been used to test models' ability to distinguish plausible and implausible events \citep{pedinotti2021did, kauf2023event}, a task that draws heavily on commonsense knowledge. The popular Winograd Schema Challenge also had a minimal pairs setup \citep{levesque2012winograd}, even though that requirement was relaxed in later versions. We here extend this approach by employing a \emph{minimal pairs-of-pairs} design, where both context and target sentences have a minimal pair counterpart.

\begin{figure*}[t!]
  \centering
  \includegraphics[width=\textwidth]{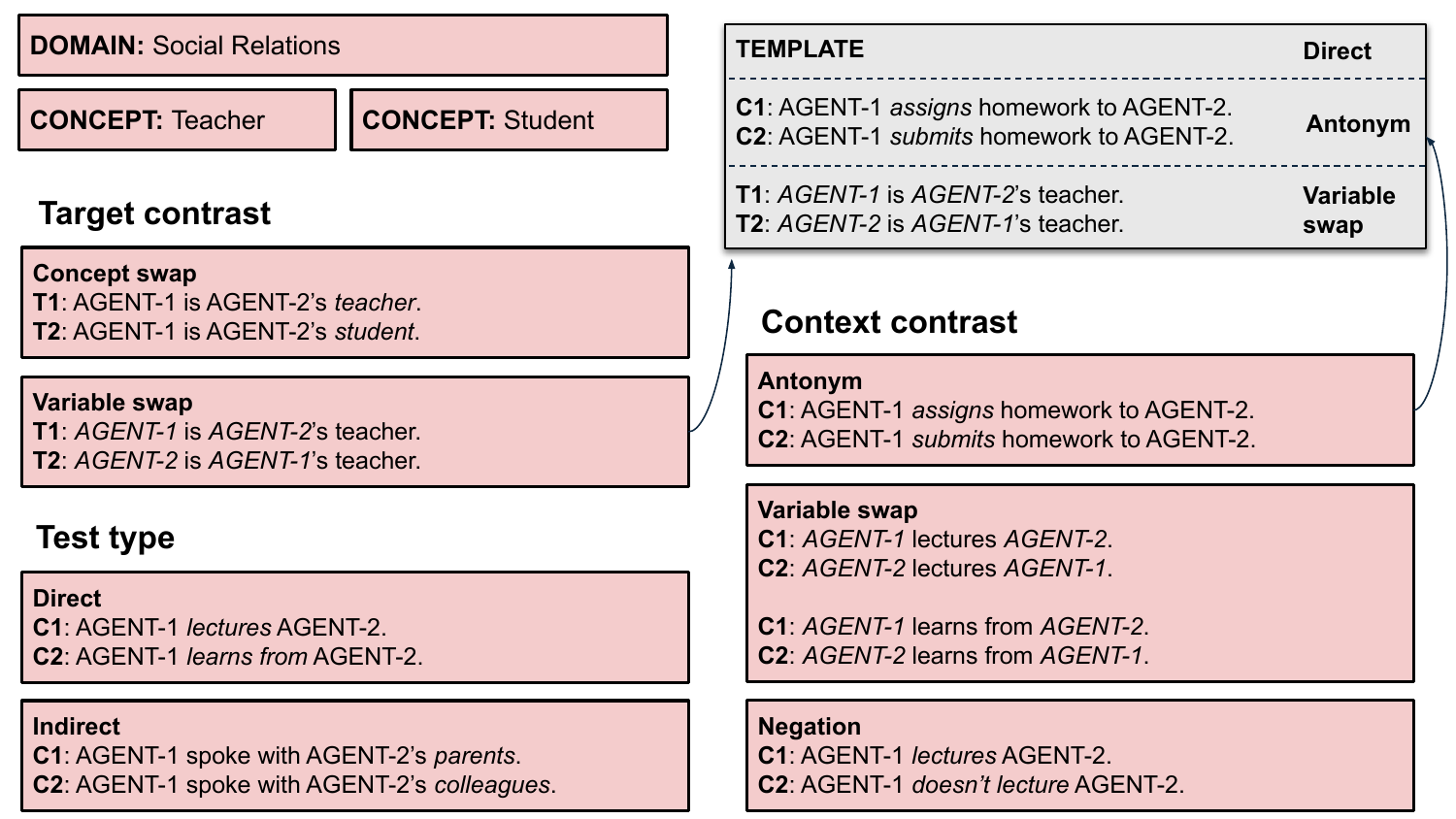}
  \caption{Types of minimal pair contrasts in Context and Target pairs. Examples shown here are for one domain and concept pair. Templates may be \emph{direct}, testing concepts explicitly, or \emph{indirect}, testing concepts implicitly using likely scenarios (e.g., a student is more likely to talk to a teacher's colleagues rather than parents). Context and target \emph{contrasts} reflect how concepts are tested. For instance, \emph{antonym} contrasts words with opposing meanings, \emph{negation} leverages ``not'', and \emph{variable swap} exploits relative ordering of entities. %
  }
  \label{fig:contrasts}
\end{figure*}

Before prompt-based evaluation became commonplace, the dominant approach for assessing language model performance on a minimal pair has been to calculate each item's (pseudo) log probability under the model. This method is effective at distinguishing grammatical and ungrammatical sentences \citep[e.g., ][]{warstadt2020blimp}, plausible and implausible events \citep{kauf2023event}, and relevant vs.~irrelevant object properties \citep{misra2023comps}, while often being calibrated to human sentence ratings and processing costs \citep{lipkin2023evaluating, shain2024large}. Yet raw log probabilities reflect a number of surface-level properties of the input, such as word frequency \citep{kauf2023event} and the number of possible paraphrases \citep{holtzman2021surface}. An alternative approach, recently made possible with more powerful LLMs, is to prompt an LLM to rate item plausibility, either absolute (on a Likert scale) or relative to the other item in the minimal pair. This approach can theoretically result in more task-specific estimates. However, for a range of linguistic and word prediction tasks, LLMs actually perform worse with direct prompting than via implicit log probability assessment \citep{hu2023prompting}, likely because of additional task demands imposed by the need to decipher instructions in the prompt \citep{hu2024auxiliary}. Thus, we report both log probability comparisons and explicit prompting results.

\section{The Framework} \label{sec:framework}

We provide a flexible generative synthetic data pipeline, capable of producing many diverse datasets, each with unique specifications and statistics, while preserving metadata and decision traces. Here, we discuss the design principles that make it possible to reuse the \ewok\ framework for systematically generating various world knowledge datasets.
In Section \ref{sec:results}, we use this framework to generate \ewokone, a systematic, broad-coverage, context-sensitive world knowledge dataset containing $4,374$ items that probe $192$ concepts (for a full concept list, see Table A\ref{tab:allconcepts}).

\paragraph{Items test concept knowledge in context.}

Effective use of world knowledge incorporates both access to a rich set of priors about the world's structure and the ability to integrate this knowledge on-the-fly with information about the current environment.
Thus, our items challenge models to leverage concepts in context.
Each item consists of two minimal pair contexts (e.g., $C_1$: \emph{The piano is in front of Ali. Ali turns \textbf{left}.}, $C_2$: \emph{The piano is in front of Ali. Ali turns \textbf{right}}) and two minimal pair target sentences (e.g., $T_1$: \emph{The piano is \textbf{right} of Ali.}, $T_2$: \emph{The piano is \textbf{left} of Ali.}). 
The two target concepts are juxtaposed such that in any item, $P(T_1 \mid C_1)>P(T_1 \mid C_2)$ and $P(T_2 \mid C_1)<P(T_2 \mid C_2)$. 
Thus, base target probabilities $P(T_1)$ and $P(T_2)$ cannot serve as plausibility cues: a model has to rely on context to establish plausibility.

\paragraph{Concepts comprise domains of world knowledge.}

\ewok\ is designed around domains of general world knowledge, with the current set of 11 domains shown in Table \ref{tab:domains}. 
We selected a range of domains that have been shown to recruit dedicated cognitive and/or neural machinery in humans, backed by extensive literature in cognitive science (introduced in Sec.~\ref{sec:para:cog-inspired}, ``Why cognition inspired?''). Domains were contributed based on past literature  during the development of \ewok\ by a team of experts in the field (authors of this paper: professors, post-docs, and grad students in cognitive science and neuroscience across a few institutions).
Each domain includes a set of concepts, also curated by the team.
For example, the domain 
\emph{social relations}
includes \emph{friend}, \emph{enemy}, \emph{teacher}, \emph{student}, \emph{boss}, \emph{subordinate}, and others.
 The number of concepts in each domain is mentioned in Table~\ref{tab:domains}.

\paragraph{Items are generated from concept-specific templates.} 
Each concept is associated with several items that test knowledge of the concept (often, but not always, by contrasting it with another concept). 
Items are generated from templates, which contain placeholders for specific objects, agents, and locations. The placeholder value should not affect the resulting plausibility judgments; thus, a template can be populated with an arbitrarily large number of fillers, which enables generating many carefully controlled items.

\paragraph{Templates consist of contrasting context and target pairs.}

A template includes 2 targets and 2 contexts. 
A \emph{target} is a simple sentence that incorporates a concept. %
A \emph{context} is a short sequence of words that can be paired with a target to yield either a plausible or an implausible combination. Context and target pairs are designed in such a way that $C_1$ but not $C_2$ matches $T_1$ and $C_2$ but not $C_1$ matches $T_2$.
The differences between the two targets or the two contexts result in specific types of contrasts, shown in Figure \ref{fig:contrasts}.

A contrasting target pair is generated using one of two mechanisms: \emph{concept swap}, which contrasts the same target with different concepts filled in, and \emph{variable swap}, which swaps two objects or agents (only possible for certain targets). 
For instance, 
{\ttfamily\ AGENT-1 is AGENT-2's teacher} can be contrasted with 
{\ttfamily AGENT-1 is AGENT-2's student} (concept swap) or with 
{\ttfamily AGENT-2 is AGENT-1's teacher} (variable swap). 

A contrasting context pair is generated using \emph{antonyms}, \emph{negation}, or \emph{variable swap}.

Finally, the templates themselves test concept knowledge in either \emph{direct} or \emph{indirect} way. Direct tests rely on immediate interpretation of a concept, whereas indirect tests leverage the probabilistic relationship between concept and target ($T_1$ doesn't have to be true given $C_1$ but it's more likely than $T_1$ given $C_2$).

\begin{figure*}[tb!]
  \centering
  \includegraphics[width=\textwidth]{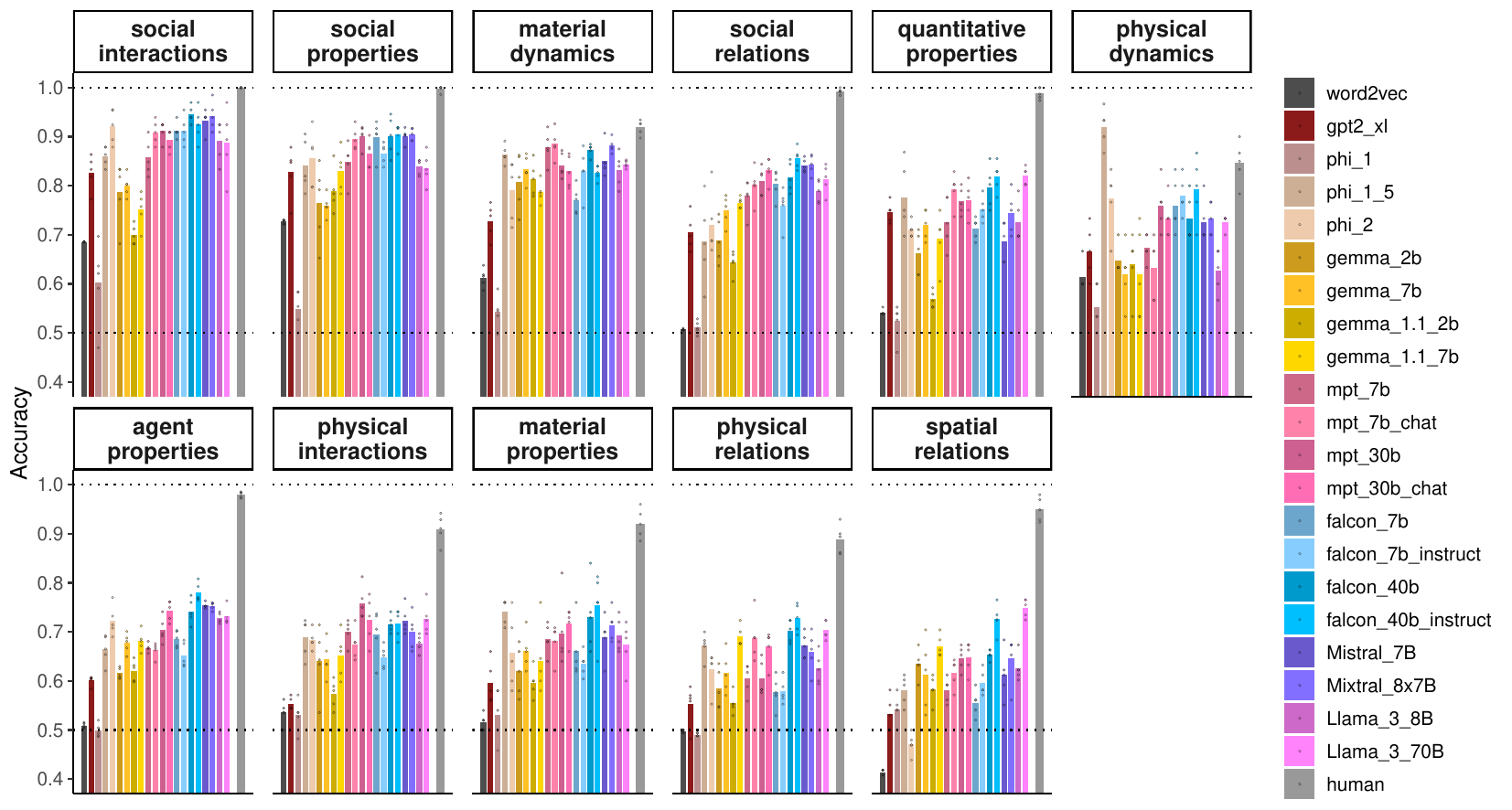}
  \caption{LLM performance across world knowledge domains (evaluated with \logprobs). Here and elsewhere, the dotted line at $0.5$ denotes chance accuracy. 
  Each dot reflects performance on a single \textit{version} of \ewokone\ (see Sec.~\ref{sec:framework} ``Templates \& fillers" and \ref{sec:dataset-generation}), with the bar reflecting the mean across the $5$  versions. LLM performance varies drastically by domain and is often substantially worse than human performance. In general, individual LLMs show similar performance patterns across domains, but these patterns are not always consistent with the human pattern.}
  \label{fig:results-bydomain}
\end{figure*}

\paragraph{Templates enable flexible yet controlled generation of items using typed fillers.}

Placeholders in templates (e.g., OBJECT or AGENT) can be restricted to only allow fillers of specific types.
For example, in the template \texttt{\{object2:can\_bounce=True\} bounced off \{object1\} from below}, 
{\ttfamily\{object2:can\_bounce=True\}} must be filled by an object marked with a flag for \texttt{can\_bounce=True}, e.g., \texttt{the ball}.
These type restrictions prevent generation of semantically anomalous or incomprehensible sentences (e.g., \texttt{The blanket bounced off the table from below)}. Type restrictions were specified by the author team during the template development stage and later validated via human behavioral ratings (Section \ref{sec:human-study} and Appendix \ref{appendix:human-data}). 

To create an initial set of object and location fillers, we generated a set of arbitrary fillers based on constraints specified in templates (e.g., objects that can bounce, that have a smell) that was filtered down to ensure the `naturalness' of the items that include these fillers. The agent fillers were selected from the list of culturally diverse high frequency male, female and non-gendered baby names. Across domains we list over $500$ filler items across $13$ classes with $28$ type restrictions. The full set of fillers is provided in the paper repo.

When populating the templates with fillers, users wanting to use the \ewok\ framework to construct their own instance of an \ewokone-like dataset can specify several parameters, e.g.: (1) the number of items to generate from each template; each full set of items is referred to as a \emph{version} (corresponding to a random seed); (2) whether fillers should be held constant across all items in a version or allowed to vary; (3) whether to apply transformations to filler restrictions at compile-time, for instance restricting all agents to take non-Western names via \texttt{agent->agent:western=False}, or swapping all objects with nonce-words via \texttt{object->nonword}. 
Such flexibility allows for controlled experimentation of the features modulating model performance.
With these arguments and others easily accessible from the command line interface, users are supported to generate many \ewok\ variations, and can additionally extend \ewok\ to add new domains, concepts, and fillers.

\section{Evaluation} \label{sec:evaluation}

Using the \ewok\ framework, we compiled the \ewokone\ dataset. To do so, our team (which includes experts in specific domains of cognitive science) curated a list of $880$
templates that test knowledge of $192$ concepts from $11$ domains. These templates were then populated with $5$ fillers that were randomly sampled from our initial filler set (described in Section \ref{sec:framework}), resulting in $5$ dataset versions. This strategy enables us to explicitly measure the variability associated with the random sampling of filler items (which, in principle, should not affect the results).
For \ewokone, we chose to hold the set of fillers constant across all items within each version, although that constraint can be relaxed if needed (see Appendix~\ref{sec:dataset-generation}).

We evaluated LLM performance on
\ewokone\ using three different approaches: traditional plausibility
estimates via querying the probability of sentences under the model, \logprobs,
as well as two prompt-based strategies, \likert\ and \choice. 
The majority of the results reported use the \logprobs\ evaluation method, which allows comparing a wide range of base and finetuned models. As we show, \logprobs\ outperforms direct prompting even for large and/or instruction-tuned models.

For the prompt-based evaluations, we collected data from both LLMs as well as human participants using paired identical prompts.
Drawing inspiration from comparative psychology, such an approach of matched evaluation has been proposed as a way to support increasingly \emph{fair} evaluations of LLMs, allowing for more direct comparison of performance, consistency, and context-sensitivity with human participants \citep{lampinen2024can}.

\subsection{Scoring metrics}
\label{sec:scoring-metrics}

For \logprobs\ evaluation, we use token-level LLM probabilities to calculate $\log P_{\theta}(T \mid C)$ as a sum of conditional log probabilities of each token:
$\sum_{k=1}^{n}  \log P_{\theta}(\mathbf{t}_{k} \mid C, \mathbf{t}_{<k})$, where $\mathbf{t}$ is the vector of tokens composing the target $T$.
For \likert, participants (humans and models) are prompted to rate the plausibility of the concatenation of each $C_i$ and $T_j$ pair on a 1--5 scale\footnote{A 1--5 point scale was chosen to distinguish between degrees of plausibility as opposed to solely absolute direction. Several items in our dataset were distinguishable through this increased granularity, e.g., a 1 vs 2 or 4 vs 5 that would not have been distinguishable otherwise.}.
This is functionally similar to the \logprobs\ condition, but explicitly based on the model's generative behavior as opposed to intrinsic to its scoring.
For \choice, participants (humans and models) are presented with $C_1$ and $C_2$, followed by a single target ($T_1$ or $T_2$), and then prompted to select the context ($1$ or $2$) that better matches the target.
This is another prompting approach that uses the language model discriminatively with access to both items.
For \likert\ and \choice, details about text generation hyperparameters can be found in \ref{appendix:text-completion} and exact prompt templates can be found in Appendix \ref{appendix:prompts}.
The metric for correctness of a given item is the recovery of the designed item structure that $\texttt{score}(T_1 \mid C_1)>\texttt{score}(T_1 \mid C_2)$ and $\texttt{score}(T_2 \mid C_1)<\texttt{score}(T_2 \mid C_2)$, where \texttt{score} reflects $P_{\theta}$ for \logprobs, the integer rating for \likert, and correct context index selection for \choice.
In all cases, models must correctly identify both $C,T$ matches to get the full score ($1.0$ point).
Identification of only one match receives $0.5$ points.
In the case of the \likert\ task, which has a coarser integer scale, if a model returns the same rating for both pairs, the model receives $0.5$ points.
Such a paradigm supports a trivial $50\%$ baseline for all scenarios.
Even a random coin flip or a deterministic model generating the same response for each query independent of context will trivially achieve this baseline.

For prompt-based evaluations (\likert\ and \choice), we used the same prompting setup across all models to assess them in the same way as humans; however, unlike humans, models were additionally provided with $2$-shot examples. These examples did not come from any of our domains and were solely meant to outline the required response formatting including one positive and one negative example. See Appendix \ref{appendix:prompts} for more details. Our initial experiments revealed that supplying these examples substantially improves model performance, providing us with an opportunity to evaluate their ``best shot'' at \ewokone.

\subsection{Models}

We evaluated $N$=20 transformer language models, selected to span a few points in the model design space.
Models primarily vary in size (\# of parameters; ranging from $1.3$B--$70$B) and pre-training diet (both \# of tokens and source of training corpora).
While most evaluated LLMs are dense pre-trained transformers ($N$=13), there are a few one-off comparisons supported including the presence of supervised fine-tuning for instructions ($N$=4) or chat ($N$=2), and mixture-of-experts (MoE) ensembling ($N$=1).
We do not intend to draw conclusions about any of these design decisions, but rather to expose variation.
Aside from these considerations in exploring variation, the selection of fine-tuned models was filtered to those that did not require specific formatting via the use of a prompt template.
Since we evaluate LLMs and humans on identical prompts, it was critical to have complete flexibility in formatting.
The full set of evaluated LLMs are listed in Fig.~\ref{fig:results-bydomain} as well as 
in Appendix~\ref{appendix:models}.

We additionally tested a baseline bag-of-words model based on {\sffamily word2vec} embeddings \citep{mikolov2013distributed}. Embeddings for each word in a context or a target were summed together to derive one vector per context/target. The context/target match was determined by a cosine similarity metric, such that an item is scored correctly if $\cos(C_1, T_1)$ > $\cos(C_2, T_1)$ (and vice versa for $T_2$). 
Word-embedding-based models provide useful baselines, and there is a long tradition of work in NLI and related tasks utilizing such easy-to-understand baselines to assess task difficulty and the presence of alternative solution strategies \cite{naik2018stress}. Past NLI work has inadvertently fallen prey to design biases, making it easy for models to rely on heuristics, such as word overlap \citep{mccoy2020right}. Additionally, word embeddings can contain rich semantic information \citep{grand2022semantic} that may help solve the task by assessing semantic relatedness between words in the context and  the target. Using bag-of-words baselines helps understand how robust the task is in the face of such heuristics. 

\subsection{Surface-level item properties}

To determine the influence of surface-level item properties on model performance, we tested whether LLM performance correlates with the number of words in an item, as well as with average word frequency in an item. Word frequency was determined using unigram counts from the Google Ngrams \citep{michel2011quantitative} 2012 American English corpus.

\begin{figure*}[t]
  \centering
  \includegraphics[width=\textwidth]{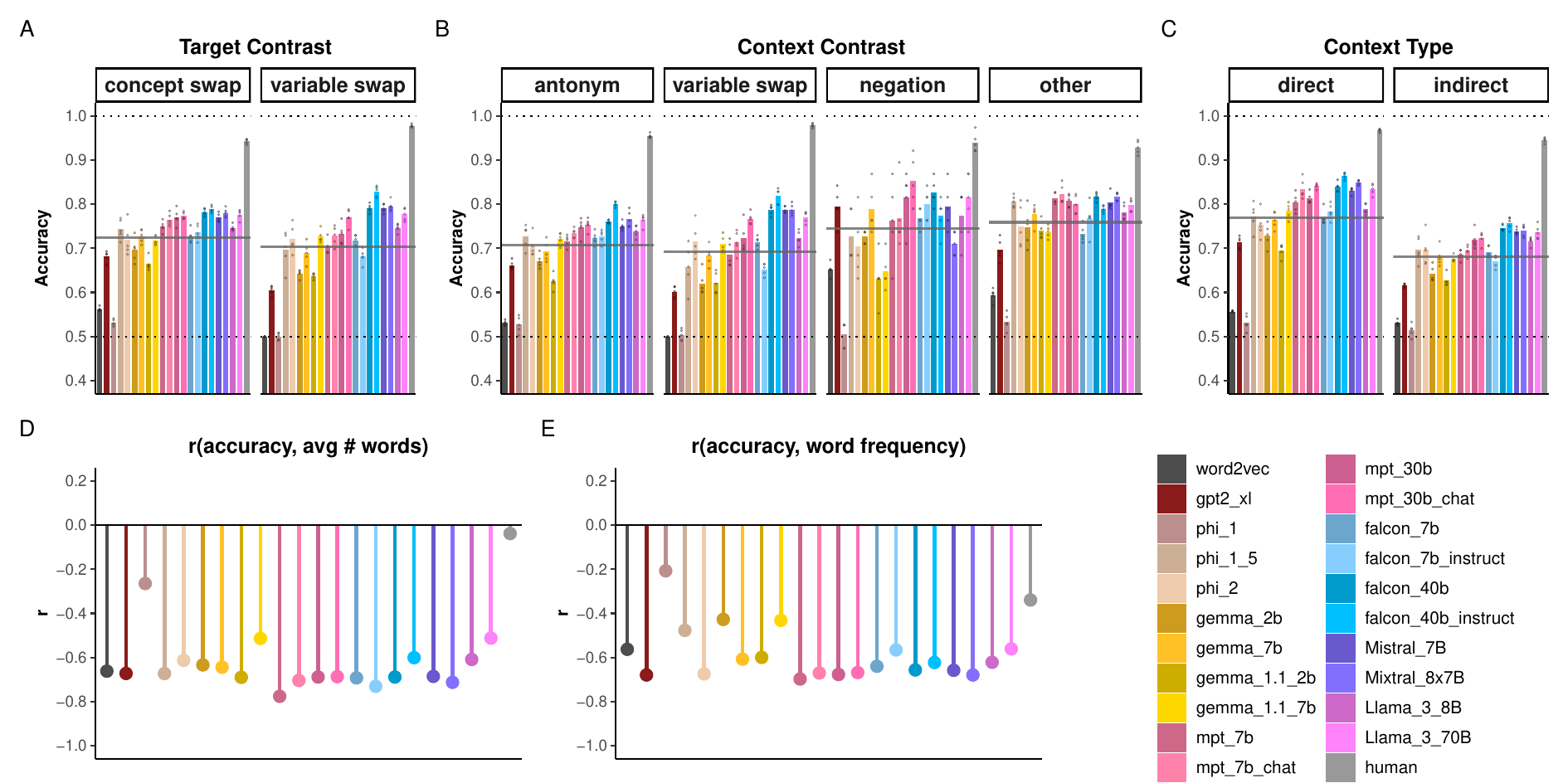}
  \caption{\emph{Top:} LLM and human performance across target contrast (A) context contrast (B) and context type (C), evaluated with \logprobs. For examples of manipulations in A-C, see Figure \ref{fig:contrasts}. Dark gray line shows average model performance. \emph{Bottom:} correlation between LLM accuracy and surface-level item features: (D) average item length and (E) average word frequency in the item. Humans are not sensitive or only weakly sensitive to these features, whereas model performance strongly correlates with them. The (counterintuitive) negative relationship between accuracy and word frequency is driven by the fact that hard domains happen to have high word frequency and is reversed once domain is controlled for (Table S\ref{tab:results-mixedmodel}).}
  \label{fig:results-design-surface-features}
\end{figure*}

\subsection{Human data}
\label{sec:human-study}

For independent norming of the items in \ewokone, we collected data from human
participants. 
 The task human participants performed was nearly identical to the \likert\ version of the task for LLMs (in a pilot study, we determined that human results for \likert\ and \choice\ show Pearson's \textit{R}=0.96 correlation, and decided to drop the \choice\ condition for the full data collection; see Appendix~\ref{appendix:human-data} (experimental design) \& Appendix~\ref{appendix:human-data-results} (results) for pilot details). 

\paragraph{Data collection}

To obtain reliable ratings across the full set of \ewokone\ items, 
we collected at least 5 responses per item from a total of {\itshape N}=1,262 participants (591 female, 579 male, 27 other/unknown; median age 36; all US-residents who reported English as their first language). Participants were recruited via Prolific ({\url{https://prolific.co}), an online study platform. %
To control for quality of data, we excluded 59 participants whose inter-subject Pearson
correlation coefficient (compared with average ratings on items from other participant who
rated the same items as this subject) was below $0.3$.
Each \ewokone\ generated item (corresponding to 5 filler-populated \textit{versions}; see Sec.~\ref{sec:framework}, ``Templates \& fillers'' and Sec.~\ref{sec:dataset-generation}) was split into four subparts: $(C_1,T_1)$, $(C_1,T_2)$, $(C_2, T_1)$, $(C_2,T_2)$ and presented in a \likert\ setting using the same prompt
as that used for LLMs (Appendix~\ref{appendix:prompts}; adapted to presentation in a web browser with a free-form text-box for human input).
Items were presented so that each participant only saw one of 5
filler-populated variants and one of the four possible sub-items $(C_i,T_j)$.
Therefore, participants provided sensibility judgments independently of any
other subparts of the item, closely matching the conditions for LLM evaluation.
Each participant rated an average of 57 items (all items came from a single domain). %
The resultant average inter-subject Pearson correlation across all items was 0.744, where the inter-subject correlation was calculated per-item based on all the participants that rated that item.

\paragraph{Aggregation and evaluation} We average ratings for a given $C_i$-$T_j$ pair across
all (typically, 5) participants that rated the pair. To obtain human norms, we
compare the averaged \likert\ ratings of subparts, awarding
$0.5$ point for each half comparison, i.e., 
$\likert(C_1,T_1) > \likert(C_2,T_1)$
and
$\likert(C_2,T_2) > \likert(C_1,T_2)$, similar to evaluating LLMs as described in Section~\ref{sec:scoring-metrics}. In case of a tie in scores for a half-comparison, we award 0 points.

\section{Experiments: \ewokone} \label{sec:results}

\paragraph{\ewokone\ can be challenging for LLMs} Although \ewokone\ was not designed to be a difficult dataset, we found that even large open models perform well below human baselines: the best model tested, {\sffamily falcon-40b-instruct}, yields a mean accuracy of $0.80$ whereas mean human accuracy is $0.95$ (Table A\ref{tab:results-mean}). As expected, larger models tend to do better, although it is not a sole predictor of performance. Instruction tuning does not consistently increase or decrease LLM performance under the \logprobs\ metric. 

\ewokone\ was later evaluated on closed frontier models via HELM \citep{liang2022holistic} in January of 2025. That evaluation was performed using \choice\ prompting given that \logprobs\ for closed models are not available. The highest performing closed models yield performance $\sim0.91$, which is still below humans ($\sim0.95$) but not by much (see Table \ref{tab:results-closedmodels}).

\begin{figure*}[tb!]
  \centering
  \includegraphics[width=\textwidth]{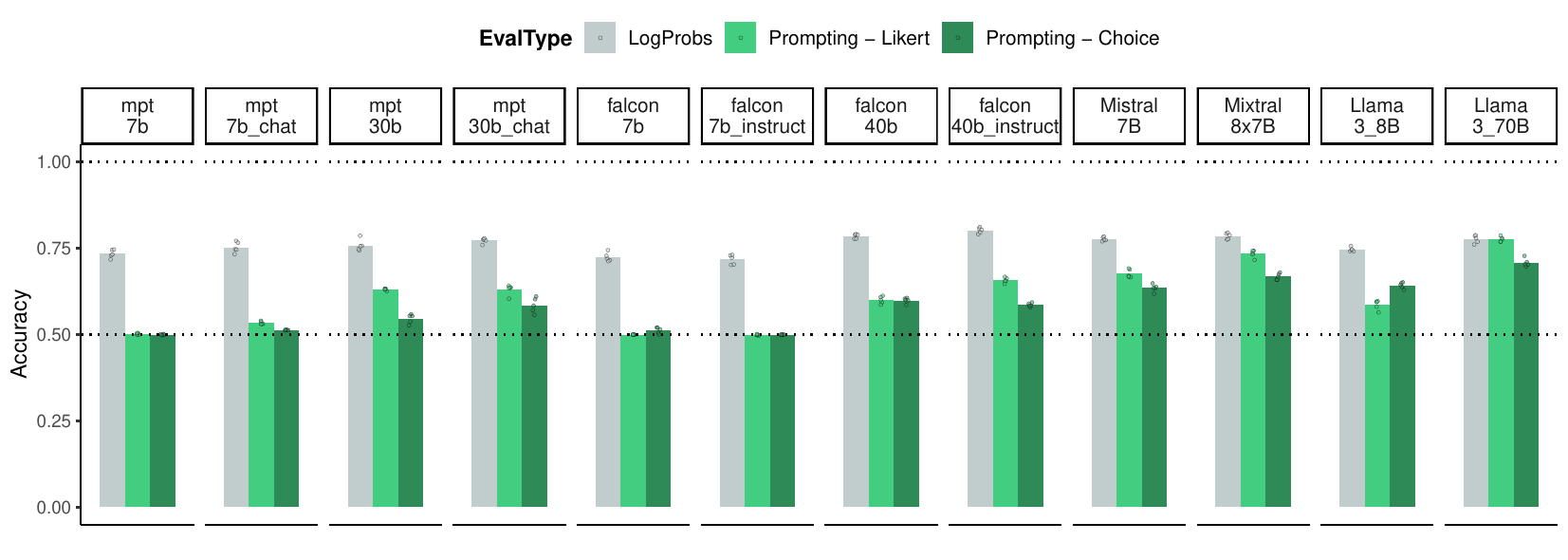}
  \caption{LLM performance assessed with \logprobs\ vs. two prompt-based tasks, \likert\ and \choice. Prompting is $2$-shot, and the outputs are constrained to the set of allowed values ($1$-$5$ for \likert, $1$ or $2$ for \choice); this setup was chosen to maximize model performance. Still, \logprobs\ is a better strategy in nearly all cases.  
  In prompting tasks, it was common for models to generate the same value, e.g., ``1'' in response to any item. Our metric was designed such that even in this scenario, the 50\% baseline would remain intact (see Section \ref{sec:framework}). Looking at \likert, without this safeguard and requiring strict inequality, the top performing {\sffamily Meta-LLama-3-70B} model drops to $59\%$. {\sffamily MPT-7B}, which performs comparably with \logprobs, drops to $2\%$ (See Table \ref{tab:results-likert-nohalf}). 
  }
  \label{fig:results-logprobs-vs-prompting}
\end{figure*}

\renewcommand{\arraystretch}{1} 
\begin{table}[ht]
\centering
\resizebox{0.5\textwidth}{!}{
\begin{tabular}{lc}
\hline
\textbf{Model} & \textbf{Mean Accuracy} \\
\hline
GPT-4 Turbo (2024-04-09) & 0.912 \\
Claude 3.5 Sonnet (2024-06-20) & 0.911 \\
Qwen2 Instruct (72B) & 0.901 \\
Claude 3 Opus (2024-02-29) & 0.893 \\
GPT-4o (2024-05-13) & 0.886 \\
Llama 3 Instruct (70B) & 0.876 \\
GPT-4o mini (2024-07-18) & 0.864 \\
Claude 3 Sonnet (2024-02-29) & 0.848 \\
Mixtral Instruct (8x22B) & 0.842 \\
Mixtral Instruct (8x7B) & 0.837 \\
Claude 3 Haiku (2024-03-07) & 0.829 \\
Claude 3.5 Haiku (2024-10-22) & 0.810 \\
GPT-3.5 Turbo (0125) & 0.797 \\
Mistral Instruct v0.3 (7B) & 0.772 \\
Llama 3 Instruct (8B) & 0.171 \\
\hline
\end{tabular}
}
\caption{Many instruction-tuned frontier LLMs perform well (but not at ceiling) in a binary choice prompt-based evaluation setting. Prompting was 2-shot (see Appendix~\ref{appendix:prompts}). The values reported here are averages over LLM performance on each domain. 
Note that \ewokone\ was made publicly available in May 2024, so models updated/released after that time may have already been exposed to it.}
\label{tab:results-closedmodels}
\end{table}

\paragraph{Performance varies drastically by domain} Figure \ref{fig:results-bydomain} shows model results by domain, from easiest to hardest (see also Table A\ref{tab:results-bydomain}). {\itshape Social interactions} is the easiest for both LLMs (mean=$0.86$, best=$0.95$) and humans ($1.0$); {\itshape spatial interactions} is both hardest for LLMs (mean=$0.62$, best=$0.75$) and shows the largest performance gap with humans (who score highly at $0.96$). We conclude that domains that tap into social knowledge are easier for LLMs whereas domains that require relational knowledge about the physical world are hardest. 

Overall, domain difficulty is consistent across LLMs, such that the same domains are predominantly hard or easy for all LLMs.
One notable exception is the heterogeneous performance of the {\sffamily phi} models, with  {\sffamily phi-1} consistently among the worst models, {\sffamily phi-1.5} outperforming all models and even humans on physical dynamics, and {\sffamily phi-2} ranging from on par with the largest models on some domains to worse than {\sffamily gpt2-xl} on {\itshape spatial relations}.
These results can perhaps be attributed to their unique training procedure, which focuses prominently on LLM-generated synthetic data.

\paragraph{LLMs show heterogeneous performance across dataset versions} 
Our generation framework is designed to allow easy substitution of different values for names, objects, and locations, among other variables. In principle, these values should not affect the results ($T_1$ still best matches $C_1$ and not $C_2$). However, models show somewhat different performance on the 5 different versions of the dataset (Table A\ref{tab:results-mean}), with {\sffamily phi-2} and {\sffamily phi-1.5} showing the largest performance range ($0.07$ for both). This heterogeneity indicates that arbitrary item choices can substantially affect model performance, such that, in our dataset and in others, ``1\%'' improvements might not be truly meaningful because they would not generalize. Humans show somewhat heterogeneous performance too (range $0.02$), although that heterogeneity is driven only by a subset of the domains (Figure \ref{fig:results-bydomain}).
One of the strengths of \ewok\ is that it allows us to explicitly explore this variation in details of the dataset construction process. In most benchmarks, these types of details in item-level construction variation are not measurable, even though these effects likely exist.

\paragraph{Domain content, item design features, and surface-level item features all affect LLM performance} 
It is critically important to test whether cross-domain performance differences are indeed driven by domain content or are rather attributable to other factors. We consider item design features (direct/indirect context; contrast type for context sentences; and contrast type for target sentences) and surface-level features (average word frequency and sentence length). Figure \ref{fig:results-design-surface-features} shows that all these factors affect model performance, often in different ways than they affect humans. %
The number of words in an item negatively affects LLM but not human performance. Unexpectedly, we found a negative relationship between word frequency and both LLM and human performance; follow-up examination of the data showed that this effect is driven by the fact that \emph{physical-relations} and \emph{spatial-relations}, the two hardest domains for LLMs, have the highest word frequency.
To evaluate the relative contributions of these factors to model performance, we jointly modeled all features using mixed effects regression. Design features and surface-level features all contributed to model performance (Table A\ref{tab:results-mixedmodel}), with word frequency having a significant positive effect and the number of words having a significant negative effect, as expected. Importantly, domain remained a significant predictor of performance even when accounting for other factors. These results indicate that surface-level factors, such as sentence length and word frequency, contribute but do not fully explain model performance.

Finally, the baseline bag-of-words model, which selects the context with the highest {\sffamily word2vec} cosine similarity to the target, performed much worse than LLMs, indicating that simple embedding similarity between words in contexts and targets is insufficient to explain LLM performance.
Of note, however, is that the {\sffamily phi-1} model does poorly on all domains and even worse than the baseline bag-of-words model. We believe the overly specific nature of phi-1's training data that consists of textbook-like demonstrations of coding problems and solutions likely makes it unable to succeed on \ewokone. 
This helps illustrate that without exposure to a naturalistic text corpus which would contain information about the world, even a model that is fluent in following English-language instructions is not able to flexibly reason about the world.

\paragraph{\logprobs\ yield higher accuracy than prompting} Evaluating LLMs with \logprobs\ resulted in above-chance accuracy for almost all open models (Figure \ref{fig:results-logprobs-vs-prompting}). As expected, the gap between \logprobs\ and prompting was larger for smaller models \citep{hu2024auxiliary}. 
While we observed this general pattern of \logprobs\ as the dominant strategy, it is possible that highly targeted prompt engineering may result in above-\logprobs\ performance.

\paragraph{Human ratings are usually, but not always, accurate} Finally, we examined discrepancies between human ratings and experimenter-designed ``ground truth'' labels. 
Sometimes, the discrepancy resulted from specific fillers changing the plausibility of a $C,T$ pair; for instance \emph{``The cooler is inside the car. Chao cannot see the cooler.''}~may be implausible because \emph{the cooler} is a large object and \emph{the car} has windows, although, for smaller objects and containers without windows, the scenario is more plausible. Sometimes, humans made mistakes. One such example is cardinal directions from the \emph{spatial relations} domain. The scenario \emph{``The bakery is north of Chao. Chao turns around. The bakery is south of Chao.''}~is implausible because cardinal directions do not depend on the agent's orientation, and yet our participants often marked it as plausible. 
Spatial cognition is known to be variable across individuals and cultures \citep{majid2004can, pitt2022different}, including preferential reliance on relative (`left'/`right') over absolute (`north'/`south') reference frames in Western cultures, which might explain why our (primarily Western) participants performed worse on items with absolute reference frames.
Overall, human data collection is a valuable source of information on our dataset but it does not replace ground truth labels.

\section{Discussion} \label{sec:discussion}

We present a systematic, flexible framework that can be used to test basic world knowledge in language models. Our goal was to develop a dataset that: 
(1) leverages a uniform item format to probe diverse domains of physical and social knowledge,
(2) presents items that employ specific concepts (``elements of world knowledge'') within these domains,
(3) requires integrating information across sentences such that the same target sentence is plausible given one context and implausible given another,
and
(4) consists of generic templates that can be used to generate a large variety of items.

We then presented evaluation results for a set of openly available models on \ewokone, a dataset generated using the \ewok\ framework. This dataset is moderately challenging for LLMs, with performance varying substantially across domains (with social knowledge being the easiest and physical and spatial knowledge being the hardest). 

We show that LLM-generated relative item probabilities, estimated with \logprobs\, allow distinguishing plausible and implausible scenarios above chance for most models and domains and that evaluation via prompting underperforms relative to \logprobs\ even for better-performing models, in agreement with prior results \citep{hu2024auxiliary,hu2023prompting,hu2024language,kauf2024comparing}. Thus, we argue that a natural way to compare models is through the probabilities they output. This is especially the case for models before post-training/instruction-tuning, weak models, or checkpoints, all of which may not have the capability of following complex instructions (e.g., multiple choice) but have substantial knowledge that can be revealed with \logprobs.

We discovered that our dataset has high \emph{domain distinguishability}---some world knowledge domains are much easier for LLMs than others---but low \emph{model distinguishability}---many LLMs perform comparably across domains. This pattern indicates that the specifics of the model architecture and training data are less important than the semantic content being evaluated. Results from all models consistently show that social knowledge is easier to learn from text alone than knowledge about the physical world. 

We also see that closed frontier models achieve high accuracy on \ewokone. This is not surprising, as this dataset was not designed as a challenge benchmark but rather as a comprehensive, broad-coverage dataset of core world knowledge in humans. Thus, the broader framework will have value even when this specific version of the dataset gets saturated.

The \ewok\ framework opens up multiple avenues for future work:
\paragraph{Targeted experiments}
The flexibility of our framework allows for conducting specific experiments using customized sets of fillers. For instance, one might investigate whether LLMs perform differently on items that include western vs.~nonwestern names, items that refer to people by names vs.~longer descriptors (``the man in the black hat''), or even items featuring nonwords (like ``\textit{florp}'') instead of real object names. 

\paragraph{Interpretability research} Knowledge editing research \citep[e.g.,][]{meng2022locating, meng2023memit} has often focused on encyclopedic knowledge; but what about knowledge of basic physical and social concepts? Our controlled minimal pair stimuli can allow researchers to identify and manipulate model circuits that might be selectively responsible for knowledge of these specific concepts across several domains.

\paragraph{From elements to world models} For a model to function as a flexible and robust general-purpose AI system, it needs to be able to construct, maintain, and update internal world models \citep{ha2018world, lecun2022path} (in cognitive science, variants of such world models are also known as mental models or situation models). The extent to which LLMs possess and use internal world models is subject to ongoing investigation \citep{hao2023reasoning, yildirim2024task, wong2023word}. The \ewok\ framework offers an opportunity to combine individual elements of world knowledge to construct multi-step scenarios for evaluating world modeling capabilities in LLMs, within and across physical and social knowledge domains.

\paragraph{Limitations} Our dataset is written in English; LLM performance might be lower on other languages, especially under-resourced ones. 
Adapting the \ewok\ framework to other languages might require redesigning the set of concepts and materials we use, which are currently grounded in the English lexicon. Thus, a multilingual framework can help more cleanly dissociate linguistic and conceptual effects on model performance.

Another limitation is that we use the same prompting setup for all models. With tailored prompt engineering, alternative generation methods, or chain-of-thought reasoning, LLM performance could improve. 

Finally, due to the synthetic nature of our dataset, some items may be atypical. 
We avoid blatant semantic violations by imposing type restrictions on template variables and validating our gold labels against human ratings, so the finalized set of items in \ewokone\ are semantically valid, just perhaps uncommon. 
There may be interest in using the \ewok\ framework to test semantically likely, not just semantically possible items \citep[e.g.,][]{hu2025shades}.  One way to achieve this would be to use LLMs to populate templates. We leave
this approach to future work, noting that LLM-based item generation should be used with care to avoid confounds that inflate model performance
\citep[e.g.,][]{panickssery2024llm}.

Another consequence of template-based generation is that items thus generated may not reflect the 
distributional statistics of natural language, and thus, log probability scores may not serve as an ideal
linking function between the knowledge language models possess and the scenarios we test them on. Methods
that explicitly identify ways to tap into LLMs' reasoning about scenario likelihood may be necessary 
\citep[e.g., identifying relevant directions in activation-space:][]{panickssery-etal-2024-steering}. %
Although we can only explore a limited set of evaluations in a single paper, the \ewok\ framework is evaluation-method-agnostic, 
and promises to continue being a useful resource allowing researchers to test for world knowledge in LLMs 
using novel methods that may be developed in the future.

\section{Release Considerations} \label{sec:release}

Our release-related goals are to (a) reduce the chances of accidental incorporation of \ewok\ items into LLM training data and (b) promote accountability and reporting when such incorporation is done intentionally. Thus:
\begin{itemize}
\item 
{\ewokone} is released on %
{HuggingFace Datasets}
\citep{lhoest2021datasets} 
with gated user access to prevent
scrapers from accessing it automatically. Users will simply accept a
\href{https://creativecommons.org/licenses/by/4.0}{CC-BY} license and
accompanying Terms of Use (ToU) wherein they will agree to explicitly report any
instances when a language model was trained on the
\ewokone\ items, and will be granted access automatically. 
\url{https://huggingface.co/datasets/ewok-core/EWoK-core-1.0}
\item The code for
the {}{\ewok\ item generation framework} is shared in a separate repository on
GitHub, with template files downloadable
as password-protected archives to prevent automatic scraping.
The repository is also protected with a ToU that requires
anyone training or fine-tuning on any data generated using \ewok\ to
report that fact. 
\url{https://github.com/ewok-core/ewok}
\item The code required to replicate the results in this
paper, along with human study and model performance data, is shared as a
separate GitHub repository following the same protections.
\url{https://github.com/ewok-core/ewok-paper}. 
\end{itemize}

\section{Conclusion}

To evaluate the ability of LLMs to construct robust world models, we need to test their ability to reason about the fundamental elements of world knowledge. The \ewok\ framework provides a way to systematically evaluate such knowledge, highlights that LLMs continue to fall short on simple scenarios requiring physical, spatial, or social knowledge, and offers opportunities for further targeted evaluations of LLMs. 

\section*{Acknowledgements}
We thank our action editor Emiel Krahmer and the anonymous reviewers for providing valuable feedback that helped improve this work.
We thank everyone who contributed ideas on dataset design, especially Hayley Ross and Yuhan Zhang, members of LINGO Lab at MIT for providing feedback on earlier versions of the work, and Yifan Mai for creating a HELM evaluation platform for \ewok. This work was supported by the Language Mission of the MIT Quest for Intelligence and by the Open Philantropy Foundation. A.I.~is additionally supported by funds from Georgia Tech School of Psychology. J.A.~is additionally supported by a Sloan Research Fellowship.

\bibliography{ewok}
\bibliographystyle{acl_natbib}
\clearpage

\appendix

\section{Methods}

\subsection{Dataset generation}
\label{sec:dataset-generation}
\ewokone\ is the result of 5 randomly-sampled filler-substituted versions of the templates in \ewok. Each {\it version} in the script below also serves as a random seed for the dataset. Each dataset includes 1 copy of each template, as specified by the {\ttfamily num\_fillers} parameter (i.e., one unique set of fillers will be sampled for this version). The parameter {\ttfamily fix\_fillers} makes it so that fillers are shared across templates within this version (i.e., {\ttfamily agent1} is always filled into by {\itshape Fatima}, if \textit{Fatima} is the first agent name sampled). Regardless of all of these parameters, variable constraints are always respected and additional fillers are sampled to satisfy constraints (e.g., {\ttfamily can\_bounce=true}).

\begin{lstlisting}[breaklines=true,frame=single,basicstyle=\ttfamily,postbreak=\mbox{{$\hookrightarrow$}\space}]
#!/bin/bash

# first, compile individual 
# concepts and corresponding 
# context/target chunks into 
# templates that can be 
# filled into with fillers

python -m ewok.compile \
    --compile_templates

# next, fill into templates using 
# custom parameters to create a 
# combined dataset

for version in {0..4}; do
    python -m ewok.compile \
    --compile_dataset=true \
    --fix_fillers=true \
    --num_fillers=1 \
    --version="$version" \
    --custom_id="ewok-core-1.0"
\end{lstlisting}

\renewcommand{\arraystretch}{0.85} %
\begin{table*}[]
  \footnotesize %
    \caption{All concepts tested in \ewokone\ grouped by domain.}
    \centering
    \begin{tabular}{
    p{2.5cm}p{12.5cm}
    }
    \toprule
        \bf Domain & \bf Concepts 
\\
\midrule
   Social Interactions & {\sf help,
deceive,
hinder,
cooperate,
compete,
evade,
seek,
chase,
learn,
teach,
respect,
insult,
flirt,
comfort,
tease,
coerce}
 \\ 
 \midrule
   Social Properties & {\sf friendly,
hostile,
trustworthy,
untrustworthy,
tolerant,
bigoted,
boastful,
humble,
dominant,
submissive,
shy,
extroverted,
introverted,
confident,
warm,
cold} 
  \\ 
 \midrule
   Social Relations & {\sf friend,
stranger,
enemy,
boss,
subordinate,
colleague,
romantic partner,
teacher,
student,
landlord,
roommate,
tenant,
parent,
sibling,
child}
 \\ 
 \midrule
   Physical Interactions & {\sf heat,
cool,
lift,
drop,
attract,
repel,
throw,
catch,
climb on,
push,
pull,
break,
fix,
collide with,
revolve around,
approach,
kick,
touch,
sit on,
stand on} 
    \\ 
 \midrule
   Physical Dynamics & {\sf roll,
slide,
fall,
rise,
sink,
float,
grow,
shrink,
oscillate,
spin,
accelerate,
slow down}
 \\ 
 \midrule
   Physical Relations & {\sf bigger,
smaller,
occlude,
occluded,
contain,
inside,
outside,
support,
supported,
attached,
touching,
connected,
block,
trail,
hang,
tied,
on,
under,
surround,
cover}
 \\ 
 \midrule
   Material Dynamics & {\sf 
fold,
ripple,
pour,
stir,
flap,
splash,
droop,
drip,
hang,
pile,
trickle,
disperse,
tap,
compress,
drape,
squeeze,
break,
rip,
wrinkle
}
  \\ 
 \midrule
   Material Properties & {\sf 
transparent,
opaque,
smooth,
rough,
heavy,
light,
glossy,
matte,
fragile,
sturdy,
cold,
warm,
soft,
hard,
elastic,
inelastic,
bouncy,
not bouncy
   }
  \\ 
 \midrule
   Agent Properties & {\sf see,
not see,
hear,
not hear,
smell,
not smell,
taste,
not taste,
feel,
not feel,
believe,
doubt,
imagine,
prefer,
have no preference,
intend,
not intend,
make an effort,
make no effort,
make more effort,
make less effort,
choose} 
  \\ 
 \midrule
   Quantitative Properties & {\sf more,
less,
fewer,
the same,
different,
a lot of,
a little,
none,
some,
all,
enough,
not enough,
many,
few,
no,
the most,
the least,
the fewest}
  \\ 
 \midrule
   Spatial Relations & {\sf left,
right,
above,
below,
north,
south,
east,
west,
in front of,
behind,
close,
far,
toward,
away} \\
       \bottomrule
    \end{tabular}

    \label{tab:allconcepts}
\end{table*}

The full list of concepts used in \ewokone\ templates is provided in Table \ref{tab:allconcepts}.

\subsection{Human data collection}
\label{appendix:human-data}

We collected human data in two phases: a pilot study used to validate our task on a single domain
and determine the measurement technique to use for data collection (\choice\ or
\likert), and a main study where we repeated data collection for the full set of
\ewokone\ items.
In the pilot study we determined that human judgments are measurement-technique-invariant \citep{ivanova2024log}, so we did not collect \choice\ judgments on the full set of materials, instead relying on  \likert\ judgments to score items.
Results from both studies are discussed in Appendix~\ref{appendix:human-data-results}.

\paragraph{Pilot study}
The pilot study was done on materials from one of the \ewokone\ subdomains---{\itshape social relations}---using one set of variables to populate the fillers (not used in the main study).
Participants from USA were recruited using Prolific, an online 
study platform, based on
being self-reported native and fluent English speakers.
We recruited a total of 30 participants across conditions. Of these, 18 reported
identifying as `female', 11 as `male', and 1 preferred not to answer. %
Participants were assigned to either the \likert\ or the
\choice\ condition, and saw items in only one of the two measurement techniques. %

Each item in \likert\ was split into four sub-items: $(C_1,T_1), (C_1,T_2), (C_2, T_1), (C_2,T_2)$. Similarly, each \choice\ item was split into two sub-items: $(C_{\{1,2\}},T_1), (C_{\{1,2\}},T_2)$.
A total of 16 participants provided \likert-scale judgments,
whereas 14 provided \choice\ responses to the items.
Most \likert\ sub-items ($C_i,T_j$) received at least 4-5 judgments, with all items receiving at least 3 judgments. The average no.~of ratings per sub-item were 4. 
All \choice\ sub-items received 7 judgments per $(C_{\{1,2\}},T_y)$ pair.
Participants never saw more than one sub-item of the same item (i.e., participants couldn't rate both $(C_1,T_1)$ and $(C_2,T_1)$ in the \likert\ study).

\paragraph{Main study}

To obtain reliable ratings across the full set of \ewokone\ items, 
we collected at least 5 responses per item from a total of {\itshape N}=1,262 participants (591 female, 579 male, 27 other/unknown; median age 36; all US residents who reported English as their first language). 
Details of the main study are reported in Section~\ref{sec:human-study}, therefore not repeated here.

\paragraph{Data exclusion}

Some online participants may respond randomly and therefore need to be excluded. To evaluate the quality of individual participants' responses, we computed Pearson R between that participant's responses and an average of other responses for that domain. Participants whose correlations were <0.3 were excluded.

\paragraph{Validating gold dataset labels using human data}

We used human data to validate author-assigned gold labels for items in \ewokone. First, we identified all items where the majority human label did not match the gold label (e.g., the plausibility of $C1$-$T2$ was higher than $C2$-$T2$). Then, we manually inspected those items to determine whether we could trace the discrepancy to an author error or to an error in the item generation pipeline. When possible, the generation pipeline was adjusted to generate the correct version of the item; alternatively, the faulty item was excluded from the dataset.

\subsection{Evaluated models}
\label{appendix:models}
The full set of evaluated models are as follows: 
{\sffamily gpt2-xl} \citep{radford2019language}, {\sffamily phi-1} \citep{gunasekar2023textbooks}, {\sffamily phi-1.5}, {\sffamily phi-2} \citep{textbooks2}, {\sffamily gemma-2b}, {\sffamily gemma-1.1-2b-it}, {\sffamily gemma-7b}, {\sffamily gemma-1.1-7b-it} \citep{team2024gemma}, {\sffamily mpt-7b}, {\sffamily mpt-7b-chat}, {\sffamily mpt-30b}, {\sffamily mpt-30b-chat} \citep{MosaicML2023Introducing30}, {\sffamily falcon-7b}, {\sffamily falcon-7b-instruct}, {\sffamily falcon-40b}, {\sffamily falcon-40b-instruct} \citep{almazrouei2023falcon}, {\sffamily mistral-7b-v0.1} \citep{jiang2023mistral}, {\sffamily mixtral-8x7b-v0.1} \citep{jiang2024mixtral}, {\sffamily Meta-Llama-3-8B}, and {\sffamily Meta-Llama-3-70B} \citep{llama3modelcard}.
All models were accessed via HuggingFace transformers \citep{wolf2020transformers}, and all experiments were run on a $4$x$A100$ $80$GB GPU cluster.

\subsection{Text completion experiments}
\label{appendix:text-completion}

For the prompting-based evaluations, \choice\ and \likert, we support two different generation options: \emph{free} and  \emph{constrained}. 
For free generation, the LLM may greedily sample up to 20 tokens, and we match the first occurrence of a valid response (a numeral between 1--2 or 1--5) with a regular expression.
Such a strategy avoids penalizing completions that begin with text or white-space, but doesn't guide the model to produce a valid response.
For constrained generation, the LLM may greedily sample from a restricted set of tokens, either 1--2 or 1--5, constrained using logit masking \citep{willard2023efficient}.
Such a strategy enforces well-structured responses, but requires a restricted response format.
In addition to variation in generation options, we support both zero- and few-shot prompting.
In Figure \ref{fig:results-logprobs-vs-prompting}, the prompting results we report use $2$-shot constrained generation, as it yields the highest performance among our space of tested strategies.

\subsection{Prompts}
\label{appendix:prompts}

For the two prompt-based evaluations, \choice\ and \likert, we include below our exact prompt templates. The \likert\ prompt was additionally used verbatim for human data evaluation.
\\\\
\choice\ Template:

\begin{lstlisting}[breaklines=true,frame=single,basicstyle=\ttfamily,postbreak=\mbox{{$\hookrightarrow$}\space}]
# INSTRUCTIONS

In this study, you will see multiple examples. In each example, you will be given two contexts and a scenario. Your task is to read the two contexts and the subsequent scenario, and pick the context that makes more sense considering the scenario that follows. The contexts will be numbered "1" or "2". You must answer using "1" or "2" in your response.

# TEST EXAMPLE

## Contexts
1. "{context1}"
2. "{context2}"

## Scenario
"{target}"

## Task
Which context makes more sense given the scenario? Please answer using either "1" or "2".

## Response
\end{lstlisting}
\phantom{.}
\\
\likert\ Template:

\begin{lstlisting}[breaklines=true,frame=single,basicstyle=\ttfamily,postbreak=\mbox{{$\hookrightarrow$}\space}]
# INSTRUCTIONS

In this study, you will see multiple examples. In each example, you will be given a scenario. Your task will be to read the scenario and answer how much it makes sense. Your response must be on a scale from 1 to 5, with 1 meaning "makes no sense", and 5 meaning "makes perfect sense".

# TEST EXAMPLE

## Scenario
"{context} {target}"

## Task
How much does this scenario make sense? Please answer using a number from 1 to 5, with 1 meaning "makes no sense", and 5 meaning "makes perfect sense".

## Response
\end{lstlisting}

\subsection{Mixed effects modeling}

To evaluate joint effects of domains, item design factors, and item surface features on LLM performance, we entered those predictors into a mixed effects logistic regression model implemented in R using \emph{lme4}. The model formula is:
\emph{
Accuracy $\sim$ 0 + Domain + ContextContrast + TargetContrast + ContextType + Frequency + NumWords + (1|Model) + (1|Item)
}

See Section \ref{sec:framework} for possible values for \emph{Domain}, \emph{ContextType}, \emph{ContextContrast}, and \emph{targetContrast}. Domain effects were estimated relative to a 0 intercept. \emph{ContextType}, \emph{ContextContrast} and \emph{TargetContrast} had deviation contrast coding. Relative word frequency and number of words per item were computed as $(C_1 + C_2)/2 + T$ (for either $T_1$ or $T_2$); these values were z-scored before being entered as regressors. \emph{Model} refers to an LLM being used, and \emph{Item} refers to each individual item, i.e.~a minimal pair of pairs. The model was fit on item-level binary accuracy data, with 1 row per target sentence. The results we report are from model performance using \logprobs\ evaluation type. See Table \ref{tab:results-mixedmodel} for results.

\section{Results}
\label{appendix:results}

\subsection{Human study}
\label{appendix:human-data-results}

\paragraph{Pilot study: Human judgments are invariant to \likert\ or \choice\ measurement}
We determined humans are closely aligned when providing \likert-scale or \choice\
judgments (Fig.~\ref{fig:likert-choice}). \likert-scale judgments are less
dependent on the specific set up (making a \choice\ judgment requires a specific
framing eliciting a comparison of two contexts given a target). In order to have
data allowing more flexible comparisons we decided to stick to \likert-scale
judgments for the full data collection in the main study.
\begin{figure}[h]
    \centering
    \includegraphics[width=\linewidth]{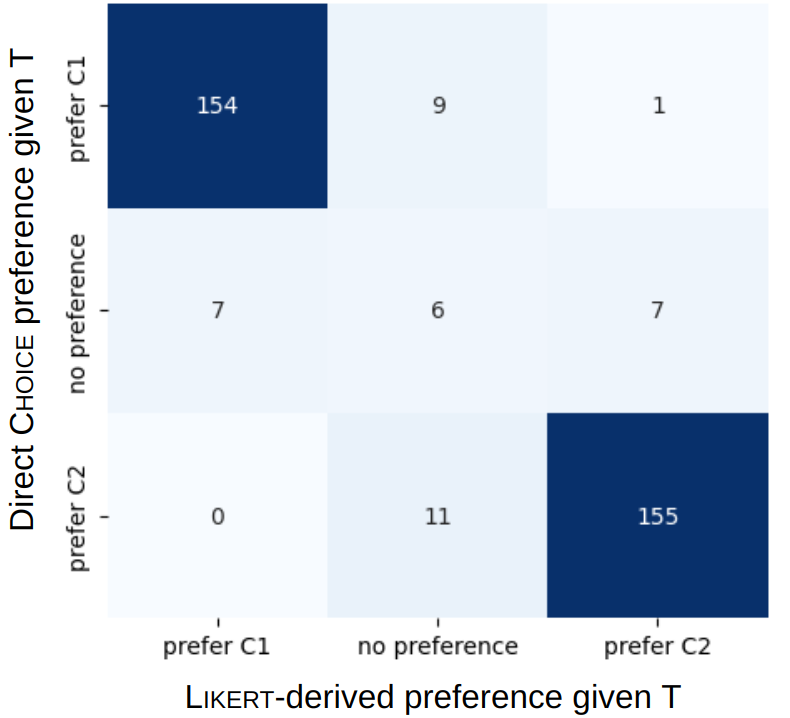}
    \caption{Confusion matrix showing agreement between a direct choice of $\{C_1,C_2\}|T_i$ and an indirect choice based on individual \likert\ ratings $\{C_i|T_i\} > \{C_j|T_i\}$.
    In this comparison we marked items
    as ``no preference'' if (1) the Likert scores weren't at least 1 apart, or (2) the average choice preference wasn't at least 0.2 away from the midpoint between two alternatives. However, we do not use a ``no preference'' bucket for the main study.
    }
    \label{fig:likert-choice}
\end{figure}

\paragraph{Full study}
See the main text Sections~\ref{sec:human-study} \& \ref{sec:results} for results about the main study (number of participants, exclusions made, inter-subject correlation, and performance of humans on the task).

\subsection{Model evaluation}

\begin{table*}[t]
\caption{Performance on \ewokone. Range reported across 5 dataset versions.\label{tab:results-mean}}
\centering
\begin{tblr}[         %
]                     %
{                     %
colspec={>{\ttfamily}Q[]Q[]Q[]},
}                     %
\hline
{\bf Model} &\bf Mean LogProbs Accuracy & \bf Range \\\hline
\textbf{human} & \textbf{0.951} & \textbf{0.942-0.957} \\
word2vec & 0.542 & 0.539-0.547 \\
gpt2\_xl & 0.655 & 0.645-0.662 \\
phi\_1 & 0.522 & 0.517-0.530 \\
phi\_1\_5 & 0.727 & 0.686-0.756 \\
phi\_2 & 0.718 & 0.696-0.771 \\
gemma\_2b & 0.678 & 0.657-0.705 \\
gemma\_7b & 0.714 & 0.697-0.734 \\
gemma\_1.1\_2b & 0.654 & 0.643-0.673 \\
gemma\_1.1\_7b & 0.720 & 0.708-0.736 \\
mpt\_7b & 0.733 & 0.716-0.745 \\
mpt\_7b\_chat & 0.751 & 0.73-0.769 \\
mpt\_30b & 0.757 & 0.743-0.786 \\
mpt\_30b\_chat & 0.771 & 0.758-0.777 \\
falcon\_7b & 0.723 & 0.713-0.744 \\
falcon\_7b\_instruct & 0.717 & 0.701-0.730 \\
falcon\_40b & 0.783 & 0.775-0.789 \\
\textbf{falcon\_40b\_instruct} & \textbf{0.801} & \textbf{0.790-0.810} \\
Mistral\_7B & 0.775 & 0.768-0.783 \\
Mixtral\_8x7B & 0.784 & 0.774-0.794 \\
Llama\_3\_8B & 0.746 & 0.740-0.756 \\
Llama\_3\_70B & 0.775 & 0.759-0.787 \\\hline
\end{tblr}
\end{table*}

\begin{table*}[t]
\caption{LLM and human performance by domain. \label{tab:results-bydomain}}
\centering
\begin{tblr}[         %
]                     %
{                     %
colspec={Q[]Q[]Q[]Q[]},
}                     %
\hline
\bf Domain & \bf LLM (average) & \bf LLM (best) & \bf Human \\ \hline %
social
interactions & 0.859 & 0.945 & 1.000 \\
social
properties & 0.839 & 0.905 & 0.997 \\
material
dynamics & 0.816 & 0.885 & 0.911 \\
social
relations & 0.761 & 0.856 & 0.992 \\
quantitative
properties & 0.725 & 0.823 & 0.986 \\
physical
dynamics & 0.706 & 0.920 & 0.833 \\
agent
properties & 0.683 & 0.778 & 0.975 \\
physical
interactions & 0.672 & 0.759 & 0.910 \\
material
properties & 0.669 & 0.755 & 0.921 \\
physical
relations & 0.627 & 0.723 & 0.886 \\
spatial
relations & 0.615 & 0.749 & 0.958 \\
\hline
\end{tblr}
\end{table*}

\begin{table*}[t]
\centering
\begin{talltblr}[         %
caption={Domain, design, and surface level features jointly contribute to LLM performance. *$p<.05$; **$p<.01$; ***$p<.001$ \label{tab:results-mixedmodel}},
]                     %
{                     %
colspec={Q[]Q[]Q[]},
}                     %
\hline
\bf Predictor Type & \bf Predictor & \bf Effect \\ \hline %
domain & social interactions & \;~1.91 *** \\
& social properties & \;~1.79 *** \\
 & material dynamics & \;~2.23 *** \\
 & social relations & \;~1.27 *** \\
 & quantitative properties & \;~1.09 *** \\
 & physical dynamics & \;~0.88 ** \\
 & agent properties & \;~0.58 ** \\
 & physical interactions & \;~0.83 *** \\
 & material properties & \;~0.75 ** \\
 & physical relations & \;~0.38  \\
 & spatial relations & \;~0.41  \\
context contrast & antonym vs.~rest & \;~0.09 *** \\
 & negation vs.~rest & \;~0.1 ** \\
 & variable swap vs.~rest & \;~0.0  \\
target contrast & variable vs.~concept swap & \;~0.0  \\
context type & direct vs.~indirect & \;~0.2 *** \\
surface features & word frequency & \;~0.07 *** \\
 & number of words & $-0.04$ ** \\
\hline \\
\end{talltblr}
\end{table*}

\begin{table*}
\centering
\begin{talltblr}[         %
caption={LLM \likert\ accuracy on \ewokone\ with stricter inequality metric. \label{tab:results-likert-nohalf}},
]                     %
{                     %
colspec={>{\ttfamily}Q[]Q[]Q[]}
}                   %
\hline
\bf Model & \bf Mean Likert Acc & \bf Range \\ 
\hline
mpt\_7b & 0.021 & 0.018-0.025  \\
mpt\_7b\_chat & 0.100 & 0.089-0.113 \\
mpt\_30b & 0.310 & 0.307-0.316 \\
mpt\_30b\_chat & 0.307 & 0.25-0.332 \\
falcon\_7b & 0.003 & 0.003-0.003 \\
falcon\_7b\_instruct & 0.005 & 0.004-0.008 \\
falcon\_40b & 0.216 & 0.185-0.24 \\
falcon\_40b\_instruct & 0.353 & 0.337-0.368 \\
Mistral\_7B & 0.396 & 0.375-0.429 \\
Mixtral\_8x7B & 0.511 & 0.471-0.533 \\
Llama\_3\_8B & 0.195 & 0.145-0.222 \\
\textbf{Llama\_3\_70B} & \textbf{0.588} & \textbf{0.576-0.603} \\
\hline
\end{talltblr}
\end{table*}

\end{document}